\documentclass[11pt,a4paper]{article}
\usepackage[hyperref]{acl2020}
\usepackage{times}
\usepackage{latexsym}

\usepackage{tablefootnote}
\usepackage{url}

\usepackage{microtype}

\aclfinalcopy 

\title{On the Universality of Deep Contextual Language Models}

\author{Shaily Bhatt\textsuperscript{1,2} \thanks{Work done while at Microsoft Research, India}, Poonam Goyal\textsuperscript{1}, Sandipan Dandapat\textsuperscript{3}, Monojit Choudhary\textsuperscript{2}, Sunayana Sitaram\textsuperscript{2}
\\
\textsuperscript{1} Birla Institute of Technology and Science, Pilani, India
\\
\textsuperscript{2} Microsoft Research, Bengaluru, India
\\
\textsuperscript{3} Microsoft R\&D, Hyderabad, India
\\
\texttt{\{f20170040, poonam\}@pilani.bits-pilani.ac.in,} \\ \texttt{ \{sadandap, monojitc, sunayana.sitaram\}@microsoft.com}}

\date{}

\begin{document}
\maketitle
\begin{abstract}
Deep Contextual Language Models (LMs) like ELMO, BERT, and their successors dominate the landscape of Natural Language Processing due to their ability to scale across multiple tasks rapidly by pre-training a single model, followed by task-specific fine-tuning. Furthermore, multilingual versions of such models like XLM-R and mBERT have given promising results in zero-shot cross-lingual transfer, potentially enabling NLP applications in many under-served and under-resourced languages. Due to this initial success, pre-trained models are being used as `Universal Language Models' as the starting point across diverse tasks, domains, and languages. This work explores the notion of `Universality' by identifying seven dimensions across which a universal model should be able to scale, that is, perform equally well or reasonably well, to be useful across diverse settings. We outline the current theoretical and empirical results that support model performance across these dimensions, along with extensions that may help address some of their current limitations. Through this survey, we lay the foundation for understanding the capabilities and limitations of massive contextual language models and help discern research gaps and directions for future work to make these LMs inclusive and fair to diverse applications, users, and linguistic phenomena.
\end{abstract}

\section{Introduction}

Language Models (LMs) have evolved considerably in the past decade, starting from the introduction of Word2Vec \cite{word2vec} to the more recent transformer-based deep models like BERT \cite{bert} and its successors. When fine-tuned with task-specific data, pre-trained LMs can be adapted to several different settings, i.e., tasks, domains, and even languages, as these LMs have been extended to multiple languages in the multilingual versions like m-BERT and derivatives. These models can be thought of as `Universal' because of their potential to be utilized `universally' in several different application scenarios.\footnote{Throughout the rest of the paper – ``these models", ``LMs", ``general domain LMs", ``contextual LMs", ``universal LMs" and all such terms refers to models including but not limited to ELMo, BERT, RoBERTa, GPT their variants, successors and multilingual versions} 

The merits of transfer learning or pre-training word representations have been known for a long time. Moreover, the recent advancements in large-scale deep learning have pushed the boundaries of intensive computation and tremendous amounts of data that can be used to pre-train LMs. However, pre-training is resource-intensive and is not carried out for specific scenarios. Instead, massive LMs are deployed into downstream applications with potentially billions of users around the world. This makes `Universality' a vital characteristic as the models must be inclusive towards a variety of language usage.

The key contributions of this paper are: 

\begin{itemize}
  \item We formally define `Universality' by selecting seven dimensions- language, multilingualism, task, domain, medium of expression, geography and demography, and time period - that capture a variety of language usage.
  \item We curate the current empirical and theoretical results that provide evidence of scaling LMs across these dimensions and identify the capabilities and gaps in these models.
   \item We outline extensions to these models that can help in overcoming current limitations to become truly universal, thus serving a larger number of end-users and scenarios.
\end{itemize}

\section{Universality and its Dimensions}

Universality can mean many things, and the associated philosophical debate is beyond the scope of this study. Our work aims not to provide a complete or exhaustive list of capabilities expected out of the model but a list of aspects that can be considered as a starting goal to achieve universality.

Our definition of universality spans seven dimensions. These are: language, multilingualism, task, domain, medium of expression, demography and geography, and time period. We selected these dimensions to cover a broad spectrum of language usage and a diverse set of NLP applications. Ideally, a truly universal model should perform strongly, or at least reasonably, across them. We present a detailed analysis of the capabilities and limitations of LMs in these dimensions in the subsequent sections. This is followed by extensions, which are techniques that can be leveraged to overcome the limitations in the particular dimension. 

\subsection{Reasoning for Selection of Dimensions}
Firstly, it is important to re-iterate that this list of seven dimensions does not intend to be an exhaustive one. Rather, these dimensions have been selected so as to cover a broad spectrum of language usage. The reasons why each of these is important for a general-purpose LM are:

\paragraph{Language} 
Massive multilingual models that can support close to 100 languages at a time are quickly becoming standard for building language technologies that cater to a wide and linguistically diverse population. However, while high-resources languages are well served, many low-resource languages are left behind \cite{joshi2021state}. Thus it is important to understand where large scale multilingual LMs stands in terms of availability, evaluation, and performance along the dimension of Language.

\paragraph{Multilingualism}
LMs are increasingly being deployed into user-facing applications and thus need to deal with real-world language usage. In multilingual (or bilingual) communities, usage of multiple languages at once gives rise to many language variations such as code-mixing, that the model will need to process. For ascertaining how well models can deal with these linguistic phenomena, understanding the capabilities and limitations of the models along this dimension becomes important.

\paragraph{Task}
LMs are increasingly becoming the standard component of most NLP pipelines. As a result, it is important to study how well they adapt to various different tasks for which they are used.

\paragraph{Domain}
Typically, LMs are trained on general purpose language, such as that obtained from Wikipedia or the Web. As such, their training data has limited signals for complex vocabulary that is common for a specialized domain such as medical, financial, legal, etc. However, real-world applications of LMs may require it to deal with information from different domains. Thus it is important to understand the limitations of these models when employed in different domain settings.

\paragraph{Medium of Expression}
Whether the language being processed is from a formal email or from an informal utterance on social media, can make significant difference in its syntactic and semantic properties. LMs being deployed in applications that span across different media are thus bound to come across linguistic variations induced due to the medium of expression. This makes it important to understand how LMs perform across different mediums of expression.

\paragraph{Geography and Demography}
Most languages in the world, including English, have multiple dialects that are influenced by geographic and demographic factors. The applications that are developed using LMs are intended to be used across the world, spanning users belonging to varied demography. It is hence important that the LMs are inclusive towards different forms of language usage and not just cater to a 'standardized' dialect of the language. Hence, it is important to understand how LMs perform across regional and social language dialects.

\paragraph{Time Period}
Given the high financial and environmental costs of training language models, a single model can be anticipated to be used for long periods of time. Language, however, changes extremely rapidly. Events happening around the world cause constant changes in the vocabulary and semantics of a language. Thus, it is important for LMs to be robust towards new word senses, sentence structures, etc. It is thus necessary to evaluate models on the dimension of Time Period as they are bound to come across language belonging to different points in the history.

The following sections go over each of these dimensions to describe the limitations and capabilities of models along each of them.

\section{Language}

There are over 7000 languages in the world. There is an increased demand for multilingual systems as information technologies penetrate more lives globally. The largest available LMs include mBERT \cite{bert}, XLM-R \cite{xlmr}, and mT5 \cite{mt5} serve 104, 100, and 101 languages respectively. It is clear that they are far from universal in terms of language coverage compared to the number of languages in the world. Further, there is an expectation that massive multilingual LMs will perform equally, or at least reasonably, well on all the languages they serve.

\begin{table}
\centering
\small
\caption{Languages and Tasks covered by different datasets and benchmarks}
\label{tab:dataset-benchmarks}
\begin{tabular}{lll} 
\hline
 \textbf{Datasets}  & \textbf{Langs.}  & \textbf{Tasks}                                               \\ 
\hline\hline
WikiANN             & 176\tablefootnote{Balanced version}                & NER                                                         \\
UD v2               & 90                & POS, dependencies\\            
XNLI                & 15                  & NLI                                                          \\
XQuAD               & 11                  & QA                                                           \\
MLDoc               & 8                   & Document classification                                      \\
MLQA                & 7                   & QA                                                           \\
PAWS-X              & 6                   & Paraphrase identification                                    \\ 
\hline
\multicolumn{3}{l}{\textbf{Benchmarks}\tablefootnote{Each task may cover only a subset of languages}}    \\ 
\hline\hline
XTREME              & 40                  & NLI, POS, NER, QA,                                           \\
                    &                     & Paraphrase identification,                                   \\
                    &                     & Sentence retrieval                                           \\
XGLUE               & 19                  & NER, POS, QA, NLI,                                           \\
                    &                     & News classification,                                         \\
                    &                     & QA matching,                                                 \\
                    &                     & Paraphrase identification,                                   \\
                    &                     & Query-ad matching,                                           \\
                    &                     & Web page ranking,                                            \\
                    &                     & Question generation,                                         \\
                    &                     & News title generation                                        \\
\hline\hline
\end{tabular}
\end{table}

The limited availability of evaluation benchmarks is a major bottleneck in knowing how  LMs perform across the languages they are pre-trained on. Table \ref{tab:dataset-benchmarks} shows that the largest benchmark, XTREME \cite{xtreme}, covers less than half of the total number of languages that these LMs are trained on. Moreover, other than datasets for syntactic tasks like NER \cite{wikiann-balanced}, and POS \cite{ud}, the largest available semantic task dataset, XNLI \cite{xnli}, covers only 15 languages. Although LMs are tested on individual tasks or languages that may not be covered in these benchmarks, overall, there is a considerable reliance on standard benchmarks to make modelling choices. Thus, how well LMs perform in the untested languages remains unanswered. 

There is a great disparity in performance across the languages that are tested through these benchmarks. A general observation is that the performance of low resource languages continues to be lower than high resource languages. The extent to which cross-lingual transfer helps improve performance varies across languages. Studies that empirically support these claims are:

Cross-script transfer is not equally good across languages in mBERT. \citealt{languagemeutralmbert} find that cross-script cross-lingual transfer is effective in the case of Hindi and Urdu, whereas this is not observed between English and Japanese. 

Word order differences across language leads to worse cross-lingual transfer \cite{languagemeutralmbert}. The correlation between word ordering distance and cross-lingual transfer is found to be high in the experiments by \citealt{ahmad2018difficulties}. \citealt{karthikeyan2019empirical} also find that word order has a significant bearing on transfer. 

Contrary to common intuition, \citealt{karthikeyan2019empirical} find that shared vocabulary does not affect Universality or generalization across languages considerably. \citealt{artetxe2019cross} also find that `effective vocabulary size per language' affects cross-lingual performance rather than joint or disjoint vocabulary of multiple languages. 

In massively multilingual LMs, where typically a joint vocabulary across languages is used, languages tend to compete for the allocations in the shared vocabulary. \citealt{siddhant2020evaluating} show that increasing number of languages may worsen performance compared to models with fewer languages. This is similar to the findings of \citealt{wu2019beto}. Thus, limiting pre-training to only the required languages needed for the downstream tasks may be more beneficial. \citealt{xlmr} coined the term \textit{``curse of multilinguality"} for this phenomenon and pointed to the trade-off between model performance and language coverage. This result is also shown in MuRIL, a BERT model trained on 17 Indian languages, which outperforms mBERT on the XTREME benchmark significantly across all languages \cite{muril}. Similarly, clustering languages and using different multilingual model for each group, rather than one massive model, gives better performance in Neural Machine Translation \cite{tan-etal-2019-multilingual}.

\citealt{wu2019beto} observe that mBERT does not transfer well between distant languages. Further, they conclude that while mBERT may perform very well in cross-lingual transfer compared to other models, it still falls short of models that have been trained with cross-lingual signals like bitext, bilingual dictionaries, or limited target language supervision. 

Answering whether all languages in mBERT are equally well represented, \citealt{wu2020all} find that mBERT does not learn high-quality representations for all languages, especially for low resource languages. The bottom 30\% languages in terms of data size perform even worse than a non-BERT model for NER. For low resource languages, the combined effect of less data for pre-training and annotated data for fine-tuning compounds together leading to worsening of their performance. On the other end of the spectrum, the top 10\% languages are hurt by joint training as mBERT performs worse than monolingual baselines of NER. 

To summarize, universality in the language dimension has three levels. At the highest level, the largest models available today span only around 100 of the 7000+ languages globally and thus are far from universal in terms of language coverage. Secondly, out of the languages that these models are trained on, not all of them are evaluated, implying that we do not have enough information to make generalized claims of universality in performance for the languages that the LMs support. Finally, at the lowest level, the performance is not uniform across the tested languages. The performance of lower resourced languages is lower than that of higher-resourced languages. Increasing the number of languages hurts performance at both ends of the spectrum, and cross-lingual transfer is non-uniform and dependent on many factors.  

\paragraph{Extensions:}
Monolingual models learn generalizable representations and can be adapted to new languages without joint training. \citealt{wu2019emerging} show that the representations learned by monolingual models without any shared vocabulary align with each other and can be adapted to a new language. Similarly, \citealt{xquad} study the transfer of monolingual representations to new languages without using shared vocabulary or joint training. They propose a zero-shot cross-lingual transfer technique where the resultant model is a monolingual LM adapted to a new language. \citealt{tigriniya} study adaptation to the extremely low resourced language, Tigrinya. They find that English XLNet generalizes better than BERT and mBERT, which is surprising given that mBERT is trained in multiple languages. Thus, the adaptation of monolingual models may help in extending LMs to new low-resource languages.

\citealt{wang2020extending} enlarge mBERT's vocabulary and continue pre-training on 27 target languages, out of which 11 are new. They observe performance improvement in zero-shot cross-lingual NER for all 27 languages. The extension benefits both the existing and newly added languages. The drawback is that the base model (mBERT) is biased towards the target languages, downgrading performance on non-target languages. 
 
The data and compute cost of training LMs from scratch poses a major limitation, especially for low-resourced languages. \citealt{englishtoforeign} propose a data and compute efficient technique to circumvent the need of training language-specific models from scratch. They learn target language word-embeddings from an English LM while keeping the pre-trained encoder layer fixed. The English and target language LMs are then both fine-tuned to obtain a bilingual LM. This technique performs better than mBERT and XLM-R on XNLI in 5 of 6 languages with different amounts of resources.
 
\citealt{chi2019can} combine the cross-lingual transfer of a multilingual LM with a task-specific monolingual LM to improve zero-shot cross-lingual classification. The source-language monolingual `teacher' model provides supervision for the downstream task, and the multilingual model acts as a `student'. The method outperforms direct multilingual fine-tuning for zero-shot cross-lingual sentiment analysis and XNLI in most of the languages.

\citealt{madx} propose an adaptor based modular framework that mitigates the curse of multilinguality and adapts a multilingual model to arbitrary tasks and languages using language and task-specific adaptors. Their method gives state-of-the-art results for cross-lingual transfer among typologically diverse languages across tasks including NER, causal commonsense reasoning, and QA.

\section{Multilingualism}

In multilingual communities, several linguistic phenomena lead to variation in language usage. While some of these are well-known and studied, others do not get enough attention. Universal LMs should be able to deal with these phenomena as we can expect them to encounter such forms of language when deployed in user-facing applications. 

Code-mixing, or using two more languages in a single utterance, is common in multilingual communities. LMs may not perform optimally in the presence of such code-mixing. Multilingual models like mBERT are not pre-trained with mixed language data, which leaves the model under-prepared for code-switched settings resulting in sub-optimal performance \cite{khanuja2020gluecos}. This can be overcome to a certain extent by training using other data, such as social media, but it is unlikely to cover all the forms of code-switching produced by multilinguals.

Romanization of languages to the Latin script has increased with the advent of digitization of communication. While some models like XLM-R \cite{xlmr} and MuRIL \cite{muril} use romanized versions of some languages in training, the effectiveness of these LMs on Romanized (or more generally transliterated) text is still unclear.

Diglossia is a kind of multilingualism where a single community uses a substantially different language dialect in different communication settings \cite{diglossia}. Since data from the internet is used in training, it is likely that LMs cannot handle diglossia. However, there are no studies that concretely prove (or disprove) this.

To summarize, apart from code-mixing, there has been very little work in recognizing, understanding, or improving LMs for different phenomena arising due to multilingualism, making the dimension under-represented in the study of LMs.

\paragraph{Extensions:} Efforts have been made to improve LMs, particularly mBERT to deal with code mixed data.~\citealt{khanuja2020gluecos} present a modified version of mBERT which performs better than standard mBERT in English-Hindi and English-Spanish code mixed data using synthetically generated code-mixed data for continued pre-training.

\section{Task}

NLP applications range from syntactic tasks, like POS, NER, etc. to semantic tasks like NLI, QA, etc. LMs learn task-agnostic representations and can be fine-tuned with task-specific data or used in task-specific architectures as features. Thus, Universal LMs should adapt well to a wide variety of tasks.

Like languages, the extent of evaluation on different NLP tasks is constrained by the availability of benchmarks that span various tasks. As shown in Table \ref{tab:dataset-benchmarks}, only a small fraction of the large number of tasks studied in NLP are evaluated by benchmarks. While there are many other task-specific datasets, the success of LMs is associated with performance on these benchmarks rather than a wider variety of tasks. 

Universal Language Model Fine-tuning for Text classification (ULMFiT) uses discriminative fine-tuning, gradual unfreezing of layers, and slanted triangular learning rates for target-specific fine-tuning, giving better performance on multiple tasks \cite{ulmfit}. This work also explicitly defines the term `universal', in their context as referring to – applicable to all tasks in text classification, using a single training process and architecture, usable without feature-engineering, and not requiring additional in-domain data. 

Masked language modeling (MLM) is the most generalizable pre-training objective for the extent of transfer among twelve pre-training objectives for nine target tasks \cite{liu2019linguistic}.

Data size is important in effective pre-training of LMs \cite{liu2019linguistic} but transfer gains between source and target tasks are also possible with smaller source datasets \cite{vu2020exploring}. 

Similarity between source and target tasks is important for transfer gains. \citealt{liu2019linguistic} find that closeness in pre-training objective and target task is important for transfer. \citealt{peters2019tune} find that while feature extraction and fine-tuning of LMs give similar performance, exceptions occur when the source and target tasks are either very similar or very dissimilar. \citealt{vu2020exploring} also find that similarity is important, especially in low-resource scenarios, but, exceptions of transfer gains between dissimilar tasks are possible.

To summarize, LMs are universal in the task dimension owing to task-specific architectures or fine-tuning. However, the success of LMs is often associated with few benchmarks which cover limited tasks. The similarity between tasks, data size, and pre-training objectives are keys to transfer gains, which are important for Universality.

\paragraph{Extensions:} 

Pattern Exploiting Training (PET) \cite{pet} reformulates tasks as cloze questions,\footnote{Cloze questions are statements with exactly one masked token.} making them the same as the MLM objective, requiring less data with no additional fine-tuning or a task-specific architecture to achieve remarkable zero-shot and few-shot performance. Using PET with ALBERT few-shot performance competitive to GPT-3, which is 780 times larger in terms of the number of parameters compared to ALBERT, is obtained  \cite{petalbert}. The recently introduced T5 model \cite{t5} leverages a text-to-text framework to enabling a single architecture to perform multiple tasks and achieving state-of-art results. These are concrete steps to enable universality towards tasks, as a single model can be built to generalize across solving multiple tasks.

\section{Domain}
NLP models are applied to many real-world applications in different domains like medical, scientific, legal, financial etc. A universal model in this dimension should adapt to different domains or scenarios without loss of performance.  

Processing domain-specific language often requires the processing of specialized vocabulary and language usage. Even though LMs learn some implicit clusters of domains \cite{aharoni2020unsupervised}, this may not be enough and specialized domain-specific LMs are needed \cite{lee2020biobert, chalkidis2020legal, beltagy2019scibert, huang2019clinicalbert, araci2019finbert, pubmedbert}.  

Despite the success of general domain LMs, it is found that pre-training LM on in-domain data improves performance across high and low resource settings \cite{gururangan2020don}.

\citealt{lee2020biobert} introduced BioBERT, learned using continued pre-training of BERT on medical text, which performed better than BERT on biomedical tasks. In contrast, \citealt{pubmedbert} introduce PubMedBERT and challenge the benefits of out-of-domain data in pre-training by showing that pre-training LM from scratch on in-domain data (if available) is better than mixed or continual pre-training. \citealt{huang2019clinicalbert} propose the clinicalBERT model that is pre-trained on clinical notes text corpora, which learns better relationships among medical concepts and outperforms general domain LMs in clinical tasks.

\citealt{chalkidis2020legal} introduce Legal-BERT and observe that continual pre-training of BERT or training it from scratch with legal data both perform similarly and significantly better than using BERT off the shelf. \citealt{beltagy2019scibert} release SciBERT, trained from scratch on scientific publications giving better performance on scientific tasks. \citealt{araci2019finbert} use domain adaptation and transfer learning to develop FinBERT, achieving state-of-the-art performance in the financial domain. 

Performance of domain-specific LM can degrade on general domain tasks \cite{pubmedbert, xu2020forget, thompson2019overcoming, rongali2020improved}. This phenomenon is also known as catastrophic forgetting and prevents the LM from being truly universal.

To summarize, LMs are not universal in the domain dimension, and different domain-specific LMs have been introduced to cater to this requirement. Domain-specific LMs are either trained from scratch or by mixed or continual pre-training of existing LMs. While none of these techniques are clear winners, performance degradation in general domain tasks is observed in many cases.

\paragraph{Extensions:}
\citealt{vu2020effective} study adversarial masking strategies to learn specific target domain vocabulary along with continual pre-training by carefully selecting tokens to be masked, leading to better domain adaptation performance across multiple source and target domains. 

MuTSPad (Multi-Task Supervised Pretraining and Adaptation) \cite{meftah2020multi} leverages hierarchical learning of a multi-task model on high-resource domain followed by fine-tuning on multiple tasks on the low-resource target domain.

\citealt{ben2020perl} extend the pivot-based transfer learning to transformer-based LMs by developing PERL (Pivot-based Encoder Representation of Language), that uses continual pre-training with MLM to learn representations that reduce the gap between source and target domains followed by fine-tuning for the downstream classification task. The pivot is selected such that the source and target domain labels have greater mutual information to facilitate a good transfer.

\citealt{jiang2007instance} propose several heuristics like removing misleading examples from the source domain, assigning more weight to target domain instances, and augmenting target training instances with predicted labels for better domain adaptation from a distributional perspective. 

Various methods that are computationally efficient \cite{poerner2020inexpensive}, use more effective adversarial training \cite{ma2019domain}, and reduce the requirement of annotated data in low-resource settings \cite{hazen2019towards} have been proposed for computation and data efficient domain adaptation.

\citealt{rongali2020improved} overcome catastrophic forgetting, out-performing domain-specific LMs while maintaining performance on general domain tasks.

\section{Medium of Expression}
Language varies with the medium of expression. There are syntactic and semantic alterations like the use of ill-formed sentence or grammatical structure, inflections, slang, compressions, and abbreviations due to limited space, familiarity with the audience, and communicative intent. Universal LMs should be robust to such variations. Often these specialized settings are simply treated as a different domain \cite{qudar2020tweetbert, nguyen2020bertweet}. However, in this work, we treat `domain' as specialized fields of expertise. The current discussion pertains to the medium in which language is used.    

Language in social media and texting may not follow conventions of written language. Sentences are often shorter or grammatically ill-formed and may not be coherent enough for LMs that rely on contextual information \cite{eisenstein2013bad,han2013lexical, choudhury2007investigation}.

LMs perform sub-optimally on non-standard language as compared to specialized LMs. BERTweet \cite{nguyen2020bertweet}, a Roberta-based LM trained on English tweets, outperforms both RoBERTa and XLM-R base models even though RoBERTa and XLM-R use 2 times and 3.75 times more data, respectively. TweetBERT \cite{qudar2020tweetbert}, another LM trained on tweets outperforms seven general domain LMs.

To summarize, the language used in different media of expression is substantially different. The limited amount of investigation in this direction indicates that LMs are not universal to these variations and perform sub-optimally on the language used in social media and texting.

\paragraph{Extensions:}
\citet{dai2020cost} propose cost-effective training by appropriate selection of additional data for training a LM from a Twitter corpus.

\section{Geography and Demography}
Standard and non-standard dialects, both social and regional, lead to varied word, and language use \cite{labov1980locating, milroy1992linguistic, tagliamonte2006analysing, wolfram2015american, sociolinguistics}. Regional dialect refers to the varied usage of the same language across different places. For example, the use of `wicked' can refer to bad or evil (``he is a wicked man"), or as an intensifier to adverbs (``my son is wicked smart") \cite{geographicsituatedlanguage, kulkarni2016freshman}. Sociolects, or social dialects, similar to regional dialects, are language dialects dependant on social variables like age, race, gender, socio-economic status, ethnicity, etc. \cite{sociolinguistics}. Geographic and demographic variations stem from grammatical, phonological, syntactic, lexical, semantic features, or any combinations, making it difficult to capture and evaluate. 

Since LMs are trained on standard language dialects, non-standard dialects spoken by millions of people are largely ignored. Such a system can result in bias against specific cultural or geographical communities in user-facing applications leading to ethical implications in building fair NLP systems. 

A Universal LM should be sensitive to semantic shifts arising from its users' demographic or geographical diversity. Taking these variations into consideration has resulted in improved performance and personalization of applications like conversational agents, sentiment analysis, word prediction, cyber-bullying detection, and machine translation \cite{ostling2016continuous, rahimi2017continuous, mirkin2015personalized, hovy2015demographic, hovy2015tagging, stoop2014using, volkova2013exploring}.

\citealt{kulkarni2016freshman} present a novel approach to quantify semantic shift that is statistically significant across geographical regions and propose a new measure of dialect similarity to establish how close the language in two regions is. \citealt{demszky2020learning} focus on Indian English and show that dialect features can be learned given very limited data with strong performance.

In the 2020 VarDial evaluation task for Romanian Dialect Identification, an SVM ensemble based on word and character n-grams outperformed fine-tuned Romanian BERT model. These results are consistent with the earlier evaluation results of VarDial where shallow models outperformed deep models. In Social Media Variety Geolocation, predicting geolocation (coordinates) from text, the best performance was obtained by a BERT architecture with a double regression classification output. In contrast, the next two best models were both shallow. \cite{gaman2020report}.

Demographic features like age, gender have been predicted from language usage \citet{peersman2011predicting, morgan2017predicting}, whereas social class or ethnicity receive less attention \cite{ardehaly2015inferring}. Prediction of demographic features from language use quantifies the correlation between social variables and social dialects. While individuals may intentionally or naturally digress from such conventions, these statistical patterns are a cornerstone for studying the interaction between society and language in computational sociolinguistics \cite{sociolinguistics}.

There are some worrisome findings of bias of performance across age, ethnicity, or gender in contextual LMs. \citep{hovy2015tagging} find that a POS tagger trained on the English Penn treebank performed better on texts written by older authors. \citealt{tan2020s} show bias against non-standard English (in this case Singaporean English) in BERT.~\citealt{bhardwaj2020investigating} expose gender bias in BERT by showing that the model assigns lower (or higher) scores consistently for sentences that contain words indicating gender in cases where gender information should have no bearing.

To summarize, the influence of geography and demography on language usage is well known, and LMs must be sensitive and inclusive of such variation. However, there has been limited, albeit now growing, attention to these factors. In some cases, shallow models have outperformed deep models in recognizing semantic shifts, and there is evidence of bias against particular social groups.

\paragraph{Extensions:}
\citealt{bamman2014distributed} learn language representations that take geographical situations or variations into account by enriching Vector space word representations (word2vec) with geographical knowledge from metadata about authors. \citealt{hovy2018capturing} employ retrofitting for including geographic information to capture regional variation in continuous regional distribution and at a fine-grained level using online posts in German and the corresponding cities of their authors as labels to create document embeddings. While these techniques do not involve any contextual LMs, such representations and retrofitting can be extended to contextual LMs. 
 
\citealt{tan2020s} propose an adversarial approach to make models like BERT more robust to non-standard forms of English using inflectional morphology perturbations. 

Debiasing techniques such as the ones studied in \citep{bolukbasi2016man, kaneko2019gender} can remove gender stereotypes from pre-trained word embeddings.

\section{Time Period}
Language evolves continuously, and individual word meanings can change significantly over the years \cite{cook2014novel, kim2014temporal, hamilton2016diachronic}. Most of the data used in LM pre-training is from the late 20th century. Thus, whether these models can handle word senses across timescales is a pertinent question. Universal LMs should appropriately deal with language with a nuanced understanding of diachronic semantic change (DSC) because, when deployed in downstream applications, such variations may be encountered and misinterpreted. Some of the studies we mention below are not strictly focused on contextual LMs. Nevertheless, we find it important to note such research as we hope it can be extended to contextual LMs in the future.

The intensity of the change of meaning of different words is different – some are more subtly changed than the others. \citealt{kim2014temporal} trace a period from 1900-2009, obtain year-specific word embeddings on the Google Books N-grams corpus, and pinpoint the extent and time-period of occurrence of semantic shift.

\citealt{hamilton2016diachronic} evaluate static word embeddings for known historical changes using corpora spanning four languages and two historical periods. They create diachronic embeddings by learning separate representations across the time periods followed by alignment over different time scales. 

DSC can be detected by clustering word sensed. \citealt{mitra2014s} organize words in a time-period specific graph where its nearest neighbors are co-occurring words, and word-senses are clustered. The shift in word sense or the emergence of a new word sense can be identified by the change of cluster for a particular word. \citealt{giulianelli2020analysing} perform clustering over usage types in BERT and use the contextual property of LM to quantify semantic change instead of relying on a specific set of word senses. 

DSC evaluation lacks standardization. \citealt{schlechtweg2019wind} perform a large scale evaluation on German, revealing the best set of parameters for optimal performance, compare various state-of-the-art methods, and outline improvements for better performance. 

Shallow models can outperform contextual LMs in identifying semantic shift. \citealt{schlechtweg2019wind} show that a shallow skip-gram model with negative sampling, orthogonal alignment, and cosine distance performs best in identifying DSC in German. \citealt{kaiser2020op} reconfirm this by using a similar model to obtain the first position in the DIACR-Ita shared task \cite{basile2020diacr} on Italian DSC. Similar findings of only limited success in contextual LMs are reported in the shared task on Unsupervised Lexical Semantic Change Detection in 5 languages hosted at SemEval 2020 \cite{schlechtweg2020semeval}.

While the success in identifying these shifts may be limited, \cite{rodina2020elmo} find that DSC identified by contextual LMs can have a strong correlation with human judgment of change. 

Within contextual LMs, BERT and ELMo perform similarly for Russian. \citealt{rodina2020elmo} show that neither BERT nor Elmo outperform each other when fine-tuned using historical text in Russian to detect semantic change. Moreover, results from the shared task on Unsupervised Lexical Semantic Change Detection in 5 languages hosted at SemEval 2020 \cite{schlechtweg2020semeval} show that systems performing well over one language may not perform as well for other languages.

To summarize, time period is under-studied and there is little understanding of whether contextual LMs can handle such nuanced language variation. For the closely related task of DSC, shallow models can outperform deep LMs, and performance can vary greatly across languages.

\paragraph{Extensions:}

\citealt{rudolph2017dynamic} develop dynamic word-embeddings with an attribute of time that captures the semantic shifts in word meanings in sequential historical data on top of Bernoulli embeddings such that representations are shared within specific time periods rather than throughout the corpus. Similarly, \citealt{bamler2017dynamic} use timestamped data to build static probabilistic representation for tracing semantic change.

To mitigate the problem of using different representations of words over different time periods, \citealt{hu2019diachronic} propose a framework for tracking and representing word senses by leveraging pre-trained BERT embeddings and Oxford dictionary data to learn fine-grained senses.   

\section{Conclusion}

Deep Contextual LMs are being applied today to various different applications due to their perceived `Universality'. In this work, we attempt to holistically define `Universality' to encompass a wide variety of scenarios and linguistic phenomena. 

We define Universality using seven dimensions: Language, Multilingualism, Task, Domain, Medium of Expression, Geography and Demography, and Time Period. These dimensions result in unique variations in language usage that are commonly encountered in real-life scenarios. We aim for this definition to be sound rather than complete. That is, a model should strive to achieve Universality in these dimensions, but they are in no way a complete, exhaustive list of everything the model needs to be capable of.

We survey research across all the dimensions and find that: First, while dimensions like language, task, and domain are more widely studied, other dimensions, especially multilingualism, geography and demography, and time period receive less attention. Second, limited evaluation benchmarks constrain the complete understanding of capabilities even in the more studied dimensions. Third, language variation arising in specific scenarios of demography, geography, time period, multilingualism, and medium of expression is often studied in an isolated manner. 

The dimensions we survey are a starting point that LMs can aim to be inclusive towards in order to serve a diverse set of users and scenarios. Large contextual LMs may not be the optimal choice for all scenarios, with shallow, task-specific models sometimes leading to better outcomes. Overall, `Universality' is yet to be fully understood, studied, and achieved. We hope that this work will lay the foundation to understanding the capabilities and limitations of LMs and spur further research into making models more inclusive and fair.

\bibliography{anthology,acl2020}
\bibliographystyle{acl_natbib}

\end{document}